\documentclass[11pt,a4paper]{article}
\usepackage[hyperref]{naaclhlt2018}
\usepackage{pslatex}
\usepackage{latexsym}

\usepackage{url}

\aclfinalcopy 
\def\aclpaperid{241} 

\usepackage{amsmath,amssymb}
\usepackage{multirow}
\usepackage{tree-dvips}
\usepackage{pgf,tikz}
\usepackage{tikz-qtree}

\usepackage{caption}

\usepackage{mathtools}
\mathtoolsset{showonlyrefs}

\usepackage[ruled,norelsize]{algorithm2e}
\SetKwComment{Comment}{$\triangleright$\ }{}

\newcommand\doc{{\it D }}
\newcommand\summary{{\it s }}
\newcommand\labels{{\it y }}
\newcommand\sentseq{{$(s_1, s_2, ..., s_n)$} }

\newcommand\labelseq{{$(y_1, y_2, ..., y_n)$} }
\newcommand\refresh{\textsc{Refresh}}

\title{Ranking Sentences for Extractive Summarization \\ with
  Reinforcement Learning}

 \author{Shashi Narayan \quad Shay B. Cohen \quad Mirella Lapata \\ 
Institute for Language, Cognition and Computation \\
 School of Informatics, University of Edinburgh \\ 
 10 Crichton Street, Edinburgh, EH8 9AB  \\ 
\url{shashi.narayan@ed.ac.uk},
\{\url{scohen,mlap}\}\url{@inf.ed.ac.uk} 
}

\date{}

\begin{document}

\maketitle

\begin{abstract}
  Single document summarization is the task of producing a shorter
  version of a document while preserving its principal information
  content.  In this paper we conceptualize extractive summarization as
  a sentence ranking task and propose a novel training algorithm which
  globally optimizes the ROUGE evaluation metric through a
  reinforcement learning objective.  We use our algorithm to train a
  neural summarization model on the CNN and DailyMail datasets and
  demonstrate experimentally that it outperforms state-of-the-art
  extractive and abstractive systems when evaluated automatically and
  by humans.\footnote{Our code and data are available here:
    \url{https://github.com/shashiongithub/Refresh}.}
\end{abstract}

\section{Introduction}
\label{sec:intro}

Automatic summarization has enjoyed wide popularity in natural
language processing due to its potential for various information
access applications. Examples include tools which aid users navigate
and digest web content (e.g.,~news, social media, product reviews),
question answering, and personalized recommendation engines. Single
document summarization --- the task of producing a shorter version of
a document while preserving its information content --- is perhaps the
most basic of summarization tasks that have been identified over the
years (see \citeauthor{Nenkova:McKeown:2011},
\citeyear{Nenkova:McKeown:2011} for a comprehensive overview).

Modern approaches to single document summarization are data-driven,
taking advantage of the success of neural network architectures and
their ability to learn continuous features without recourse to
preprocessing tools or linguistic annotations.  \emph{Abstractive}
summarization involves various text rewriting operations
(e.g.,~substitution, deletion, reordering) and has been recently
framed as a sequence-to-sequence problem
\cite{sutskever-nips14}. Central in most approaches
\cite{rush-acl15,chenIjcai-16,nallapati-signll16,see-acl17,tanwan-acl17,paulus-socher-arxiv17}
is an encoder-decoder architecture modeled by recurrent neural
networks. The encoder reads the source sequence into a list of
continuous-space representations from which the decoder generates the
target sequence. An attention mechanism \cite{bahdanau-arxiv14} is
often used to locate the region of focus during decoding.

\emph{Extractive} systems create a summary by identifying (and
subsequently concatenating) the most important sentences in a
document. A few recent approaches
\cite{jp-acl16,nallapati17,narayan-arxiv17,Yasunaga:2017:gcn}
conceptualize extractive summarization as a sequence labeling task in
which each label specifies whether each document sentence should be
included in the summary. Existing models rely on recurrent neural
networks to derive a meaning representation of the document which is
then used to label each sentence, taking the previously labeled
sentences into account. These models are typically trained using
cross-entropy loss in order to maximize the likelihood of the
ground-truth labels and do not necessarily \emph{learn to rank}
sentences based on their importance due to the absence of a
ranking-based objective. Another discrepancy comes from the mismatch
between the learning objective and the evaluation criterion, namely
ROUGE \cite{rouge}, which takes the entire summary into account.

In this paper we argue that cross-entropy training is not optimal for
extractive summarization. Models trained this way are prone to
generating verbose summaries with unnecessarily long sentences and
redundant information.  We propose to overcome these difficulties by
globally optimizing the ROUGE evaluation metric and learning to rank
sentences for summary generation through a reinforcement learning
objective. Similar to previous work
\cite{jp-acl16,narayan-arxiv17,nallapati17}, our neural summarization
model consists of a hierarchical document encoder and a hierarchical
sentence extractor. During training, it combines the
maximum-likelihood cross-entropy loss with rewards from policy
gradient reinforcement learning to directly optimize the evaluation
metric relevant for the summarization task. We show that this global
optimization framework renders extractive models better at
discriminating among sentences for the final summary; a sentence is
ranked high for selection if it often occurs in high scoring
summaries.

We report results on the CNN and DailyMail news highlights datasets
\cite{hermann-nips15} which have been recently used as testbeds for
the evaluation of neural summarization systems. Experimental results
show that when evaluated automatically (in terms of ROUGE), our model
outperforms state-of-the-art extractive \emph{and} abstractive
systems. We also conduct two human evaluations in order to assess
(a)~which type of summary participants prefer (we compare extractive
and abstractive systems) and (b)~how much key information from the
document is preserved in the summary (we ask participants to answer
questions pertaining to the content in the document by reading system
summaries). Both evaluations overwhelmingly show that human subjects
find our summaries more informative and complete.

Our contributions in this work are three-fold: a novel application
of reinforcement learning to sentence ranking for extractive
summarization; corroborated by analysis and empirical results
showing that cross-entropy training is not well-suited to the
summarization task; and large scale user studies following two
evaluation paradigms which demonstrate that state-of-the-art
abstractive systems lag behind extractive ones when the latter are
globally trained.

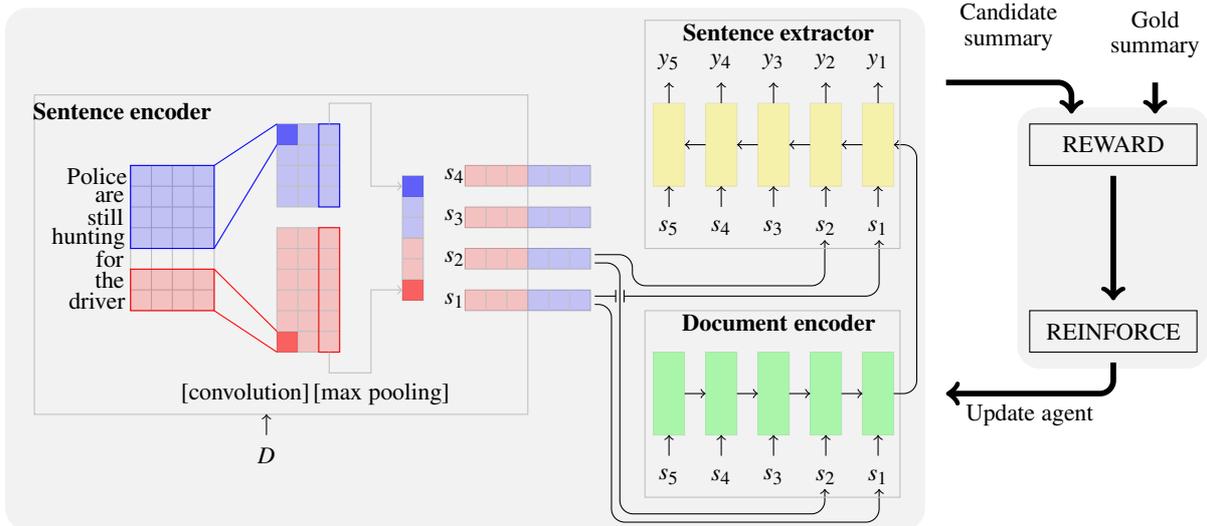
\begin{figure*}[t!]
  \center{\footnotesize
    \begin{tikzpicture}[scale=0.55]
      
      \fill [gray,opacity=0.1,rounded corners=3mm] (21.2,-1.4) rectangle (25.8,4.9);

      \draw [rounded corners=5pt,solid,->,line width=0.75mm]
      (19.5,5.5) -- (22.5, 5.5) node
            [above,midway,label={[align=center]above:Candidate\\summary}]
            {} -- (22.5, 4.75) ;

      \draw [rounded corners=5pt,solid,->,line width=0.75mm] (24.5,
      5.5) -- (24.5, 4.75) node [above,near
        start,label={[align=center]above:Gold\\summary}] {} ;

      \draw (21.5,3.5) rectangle (25.5,4.5);
      \node at (23.5,4) {REWARD}; 

      \draw [rounded corners=5pt,solid,->,line width=0.75mm]
      (23.5,3.25) -- (23.5, 0.25);

      \draw (21.5,-1) rectangle (25.5,0);
      \node at (23.5,-0.5) {REINFORCE};  

      \draw [rounded corners=5pt,solid,->,line width=0.75mm]
      (23.5,-1.25) -- (23.5, -2) -- (19.5, -2) node [below,midway]
            {Update agent};

      \fill [gray,opacity=0.1,rounded corners=3mm] (-3,-5.3) rectangle (19,7.3);
      
      \begin{scope}[shift={(0,0)}] 
        \draw [lightgray] (0,0) -- (0, 3.5) -- (2,3.5) -- (2,0) -- (0,0);
        \draw [lightgray] (0,0.5) -- (2, 0.5); \draw [lightgray] (0,1) -- (2, 1);
        \draw [lightgray] (0,1.5) -- (2, 1.5); \draw [lightgray] (0,2) -- (2, 2);
        \draw [lightgray] (0,2.5) -- (2, 2.5); \draw [lightgray] (0,3) -- (2, 3); 
        \draw [lightgray] (0.5,0) -- (0.5, 3.5); \draw [lightgray] (1,0) -- (1, 3.5);
        \draw [lightgray] (1.5,0) -- (1.5, 3.5); 
        \node at (-0.8,3.25) {Police};
        \node at (-0.5,2.75) {are};
        \node at (-0.6,2.25) {still};
        \node at (-1,1.75) {hunting};
        \node at (-0.5,1.25) {for};
        \node at (-0.5,0.75) {the};
        \node at (-0.8,0.25) {driver};
        
        \fill [red,opacity=0.2] (0,0) rectangle (2,1);
        \draw [red] (0,0) rectangle (2,1);
        \fill [red,opacity=0.2] (3.5,-1) rectangle (5,2);
        \draw [lightgray] (3.5,-1) rectangle (5,2);
        \draw [lightgray] (3.5,-0.5) -- (5, -0.5); \draw [lightgray] (3.5,0) -- (5, 0);
        \draw [lightgray] (3.5, 0.5) -- (5, 0.5); \draw [lightgray] (3.5,1) -- (5, 1);
        \draw [lightgray] (3.5,1.5) -- (5, 1.5); 
        \draw [lightgray] (4,-1) -- (4, 2); \draw [lightgray] (4.5,-1) -- (4.5, 2);
        \fill [red,opacity=0.5] (3.5,-1) rectangle (4,-0.5);
        \draw [red] (2,0) -- (3.5,-1); \draw [red] (2,1) -- (3.5,-0.5); 
        
        \fill [blue,opacity=0.2] (0,1.5) rectangle (2,3.5);
        \draw [blue] (0,1.5) rectangle (2,3.5);
        \fill [blue,opacity=0.2] (3.5,2.5) rectangle (5,4.5);
        \draw [lightgray] (3.5,2.5) rectangle (5,4.5);
        \draw [lightgray] (3.5,3) -- (5, 3); \draw [lightgray] (3.5,3.5) -- (5, 3.5);
        \draw [lightgray] (3.5, 4) -- (5, 4); 
        \draw [lightgray] (4,2.5) -- (4, 4.5); \draw [lightgray] (4.5,2.5) -- (4.5, 4.5);
        \fill [blue,opacity=0.5] (3.5,4) rectangle (4,4.5);
        \draw [blue] (2,3.5) -- (3.5,4.5); \draw [blue] (2,1.5) -- (3.5,4); 
        
        \fill [blue,opacity=0.2] (6.5,3.25) rectangle (7,1.75);
        \draw [lightgray] (6.5,3.25) rectangle (7,1.75);
        \draw [lightgray] (6.5,2.75) -- (7, 2.75); \draw [lightgray] (6.5,2.25) -- (7, 2.25);
        \fill [blue,opacity=0.5] (6.5,2.75) rectangle (7,3.25);
        \fill [red,opacity=0.2] (6.5,1.75) rectangle (7,0.25);
        \draw [lightgray] (6.5,1.75) rectangle (7,0.25);
        \draw [lightgray] (6.5,1.25) -- (7, 1.25); \draw [lightgray] (6.5,0.75) -- (7, 0.75);
        \fill [red,opacity=0.5] (6.5,0.75) rectangle (7,0.25);
        
        \draw [blue] (4.5,2.5) rectangle (5,4.5);
        \draw [lightgray,->] (4.75,4.5) -- (4.75, 5) -- (5.75, 5) --
        (5.75, 3) -- (6.5, 3);
        \draw [red] (4.5,-1) rectangle (5,2);
        \draw [lightgray,->] (4.75,-1) -- (4.75, -1.5) -- (5.75, -1.5)
        -- (5.75, 0.5) -- (6.5, 0.5);
        
        \node at (2.75,-2) {[convolution]};
        \node at (6,-2) {[max pooling]};

        \draw [lightgray] (-2.3,-2.5) rectangle (9.5, 5.2);
        \node at (-0.2,4.8) {\textbf{Sentence encoder}};        
        \draw [->] (3.25, -3) -- (3.25, -2.5) ;
        \node at (3.25,-3.5) {\doc};

        \draw [lightgray] (8,3) rectangle (11,3.5);
        \fill [red,opacity=0.2] (8,3) rectangle (9.5,3.5);
        \fill [blue,opacity=0.2] (9.5,3) rectangle (11,3.5);
        \draw [lightgray] (8.5,3) -- (8.5,3.5);
        \draw [lightgray] (9,3) -- (9,3.5);
        \draw [lightgray] (9.5,3) -- (9.5,3.5);
        \draw [lightgray] (10,3) -- (10,3.5);
        \draw [lightgray] (10.5,3) -- (10.5,3.5);
        \node at (7.75,3.25) {$s_4$};
                
        \draw [lightgray] (8,2) rectangle (11,2.5);
        \fill [red,opacity=0.2] (8,2) rectangle (9.5,2.5);
        \fill [blue,opacity=0.2] (9.5,2) rectangle (11,2.5);
        \draw [lightgray] (8.5,2) -- (8.5,2.5);
        \draw [lightgray] (9,2) -- (9,2.5);
        \draw [lightgray] (9.5,2) -- (9.5,2.5);
        \draw [lightgray] (10,2) -- (10,2.5);
        \draw [lightgray] (10.5,2) -- (10.5,2.5);
        \node at (7.75,2.25) {$s_3$};

        \draw [lightgray] (8,1) rectangle (11,1.5);
        \fill [red,opacity=0.2] (8,1) rectangle (9.5,1.5);
        \fill [blue,opacity=0.2] (9.5,1) rectangle (11,1.5);
        \draw [lightgray] (8.5,1) -- (8.5,1.5);
        \draw [lightgray] (9,1) -- (9,1.5);
        \draw [lightgray] (9.5,1) -- (9.5,1.5);
        \draw [lightgray] (10,1) -- (10,1.5);
        \draw [lightgray] (10.5,1) -- (10.5,1.5);
        \node at (7.75,1.25) {$s_2$};
        
        \draw [rounded corners=5pt,->] (11.1, 1.35) -- (12, 1.35) --
        (12, 0.6) -- (16.6, 0.6) -- (16.6, 1.7);
        
        \draw [rounded corners=5pt,->] (11.1, 1.15) -- (11.7, 1.15) --
        (11.7, -4.9) -- (16.6, -4.9) -- (16.6, -4.3);

        \draw [lightgray] (8,0) rectangle (11,0.5);
        \fill [red,opacity=0.2] (8,0) rectangle (9.5,0.5);
        \fill [blue,opacity=0.2] (9.5,0) rectangle (11,0.5);
        \draw [lightgray] (8.5,0) -- (8.5,0.5);
        \draw [lightgray] (9,0) -- (9,0.5);
        \draw [lightgray] (9.5,0) -- (9.5,0.5);
        \draw [lightgray] (10,0) -- (10,0.5);
        \draw [lightgray] (10.5,0) -- (10.5,0.5);
        \node at (7.75,0.25) {$s_1$};

        \draw (11.1, 0.35) -- (11.6, 0.35); 
        \draw (11.6, 0.2) -- (11.6, 0.5);
        \draw (11.8, 0.2) -- (11.8, 0.5);
        \draw [rounded corners=5pt,->] (11.8, 0.35) -- (17.9, 0.35) -- (17.9, 1.7);
        
        \draw [rounded corners=5pt,->] (11.1, 0.15) -- (11.5, 0.15) --
        (11.5, -5.1) -- (17.9, -5.1) -- (17.9, -4.3);

                
      \end{scope}

      \begin{scope}[shift={(12.5,-3)}] 
        \draw [lightgray] (-0.2,-1.5) rectangle (5.9, 3);
        \node at (3,2.7) {\textbf{Document encoder}};

        \draw [lightgray,fill=green,opacity=0.3] (0,0) rectangle (0.75,2);
        \draw [lightgray,fill=green,opacity=0.3] (1.25,0) rectangle (2,2);
        \draw [lightgray,fill=green,opacity=0.3] (2.50,0) rectangle (3.25,2);
        \draw [lightgray,fill=green,opacity=0.3] (3.75,0) rectangle (4.5,2);
        \draw [lightgray,fill=green,opacity=0.3] (5,0) rectangle (5.75,2);
        
        \draw [->] (0.375, -0.5) -- (0.375, 0) ;
        \node at (0.375,-1) {$s_5$};
        \draw [->] (1.625, -0.5) -- (1.625, 0) ;
        \node at (1.625,-1) {$s_4$};
        \draw [->] (2.875, -0.5) -- (2.875, 0) ;
        \node at (2.875,-1) {$s_3$};
        \draw [->] (4.125, -0.5) -- (4.125, 0) ;
        \node at (4.125,-1) {$s_2$};
        \draw [->] (5.375, -0.5) -- (5.375, 0) ;
        \node at (5.375,-1) {$s_1$};

        \draw [->] (0.75, 1) -- (1.25, 1) ;
        \draw [->] (2, 1) -- (2.50, 1) ;
        \draw [->] (3.25, 1) -- (3.75, 1) ;
        \draw [->] (4.5, 1) -- (5, 1) ;

        \draw [rounded corners=5pt,->] (5.75, 1) -- (6.3, 1) --
        (6.3,7) -- (5.75,7);
      \end{scope}
      
      \begin{scope}[shift={(12.5,3)}]
        \draw [lightgray] (-0.2,-1.5) rectangle (5.9, 4);
        \node at (3,3.7) {\textbf{Sentence extractor}};

        \draw [lightgray,fill=yellow,opacity=0.3] (0,0) rectangle (0.75,2);
        \draw [lightgray,fill=yellow,opacity=0.3] (1.25,0) rectangle (2,2);
        \draw [lightgray,fill=yellow,opacity=0.3] (2.50,0) rectangle (3.25,2);
        \draw [lightgray,fill=yellow,opacity=0.3] (3.75,0) rectangle (4.5,2);
        \draw [lightgray,fill=yellow,opacity=0.3] (5,0) rectangle (5.75,2);
        
        \draw [->] (0.375, -0.5) -- (0.375, 0) ;
        \node at (0.375,-1) {$s_5$};
        \draw [->] (1.625, -0.5) -- (1.625, 0) ;
        \node at (1.625,-1) {$s_4$};
        \draw [->] (2.875, -0.5) -- (2.875, 0) ;
        \node at (2.875,-1) {$s_3$};
        \draw [->] (4.125, -0.5) -- (4.125, 0) ;
        \node at (4.125,-1) {$s_2$};
        \draw [->] (5.375, -0.5) -- (5.375, 0) ;
        \node at (5.375,-1) {$s_1$};

        \draw [<-] (0.75, 1) -- (1.25, 1) ;
        \draw [<-] (2, 1) -- (2.50, 1) ;
        \draw [<-] (3.25, 1) -- (3.75, 1) ;
        \draw [<-] (4.5, 1) -- (5, 1) ;
        
        \draw [->] (0.375, 2) -- (0.375, 2.50) ;
        \node at (0.375,3) {$y_5$};
        \draw [->] (1.625, 2) -- (1.625, 2.50) ;
        \node at (1.625,3) {$y_4$};
        \draw [->] (2.875, 2) -- (2.875, 2.50) ;
        \node at (2.875,3) {$y_3$};
        \draw [->] (4.125, 2) -- (4.125, 2.50) ;
        \node at (4.125,3) {$y_2$};
        \draw [->] (5.375, 2) -- (5.375, 2.50) ;
        \node at (5.375,3) {$y_1$};
      \end{scope}
     
    \end{tikzpicture}
  }
  \caption{Extractive summarization model with reinforcement learning:
    a hierarchical encoder-decoder model ranks sentences for their
    extract-worthiness and a candidate summary is assembled from the
    top ranked sentences; the REWARD generator compares the candidate
    against the gold summary to give a reward which is used in the
    REINFORCE algorithm \protect\cite{Williams:1992} to update the
    model.} \label{fig:architecture}
\end{figure*}

\section{Summarization as Sentence Ranking}
\label{sec:extr-summ-as}




Given a document \doc consisting of a sequence of sentences \sentseq, an
extractive summarizer aims to produce a summary $\mathcal{S}$ by
selecting~$m$ sentences from \doc (where $m < n$). For each sentence
\mbox{$s_i \in \doc$}, we predict a label $y_i \in \{0,1\}$ (where~$1$
means that $s_i$ should be included in the summary) and assign a score
$p(y_i| s_i,\doc,\theta)$ quantifying $s_i$'s relevance to the
summary.  The model learns to assign $p(1| s_i,\doc,\theta) > p(1|
s_j,\doc,\theta)$ when sentence~$s_i$ is more relevant
than~$s_j$. Model parameters are denoted by $\theta$.  We
estimate~$p(y_i| s_i,\doc,\theta)$ using a neural network model and
assemble a summary $\mathcal{S}$ by selecting~$m$ sentences with top
$p(1| s_i,\doc,\theta)$ scores.

Our architecture resembles those previously proposed in the literature
\cite{jp-acl16,nallapati17,narayan-arxiv17}.  The main components
include a sentence encoder, a document encoder, and a sentence
extractor (see the left block of Figure~\ref{fig:architecture}) which
we describe in more detail below.




\paragraph{Sentence Encoder}
A core component of our model is a convolutional sentence encoder
which encodes sentences into continuous representations. In recent
years, CNNs have proven useful for various NLP tasks
\cite{nlpscratch,kim-emnlp14,kalchbrenner-acl14,zhang-nips15,lei-emnlp15,kim-aaai16,jp-acl16}
because of their effectiveness in identifying salient patterns in the
input \cite{showattendtell}. In the case of summarization, CNNs can
identify named-entities and events that correlate with the gold
summary.

We use temporal narrow convolution by applying a kernel filter~$K$ of
width~$h$ to a window of~$h$ words in sentence $s$ to produce a new
feature. This filter is applied to each possible window of words
in~$s$ to produce a feature map $f \in R^{k-h+1}$ where $k$ is the
sentence length. We then apply max-pooling over time over the feature
map~$f$ and take the maximum value as the feature corresponding to
this particular filter~$K$. We use multiple kernels of various sizes
and each kernel multiple times to construct the representation of a
sentence.
In Figure~\ref{fig:architecture}, kernels of size~$2$ (red) and~$4$
(blue) are applied three times each. Max-pooling over time yields two
feature lists $f^{K_{2}}$ and $f^{K_{4}} \in R^3$. The final sentence
embeddings have six dimensions.

\paragraph{Document Encoder}

The document encoder composes a sequence of sentences to obtain a
document representation. We use a recurrent neural network with Long
Short-Term Memory (LSTM) cells to avoid the vanishing gradient problem
when training long sequences \cite{lstm}. Given a document \doc
consisting of a sequence of sentences $(s_1, s_2, \ldots , s_n)$, we
follow common practice and feed sentences in reverse order
\cite{sutskever-nips14,lijurafsky-acl15,katja-emnlp15,narayan-arxiv17}. This
way we make sure that the network also considers the top sentences of
the document which are particularly important for summarization
\cite{rush-acl15,nallapati-signll16}.

\paragraph{Sentence Extractor}

Our sentence extractor sequentially labels each sentence in a document
with~$1$ (relevant for the summary) or~$0$ (otherwise). It is
implemented with another RNN with LSTM cells and a softmax layer. At
time $t_i$, it reads sentence $s_i$ and makes a binary prediction,
conditioned on the document representation (obtained from the document
encoder) and the previously labeled sentences. This way, the sentence
extractor is able to identify locally and globally important sentences
within the document. We rank the sentences in a document~\doc by
\mbox{$p(y_i=1 | s_i,\doc,\theta)$}, the confidence scores assigned by
the softmax layer of the sentence extractor.




We learn to rank sentences by training our network in a reinforcement
learning framework, directly optimizing the final evaluation metric,
namely ROUGE \cite{rouge}. Before we describe our training algorithm,
we elaborate on why the maximum-likelihood cross-entropy objective
could be deficient for ranking sentences for summarization
(Section~\ref{sec:crossent}). Then, we define our reinforcement
learning objective in Section~\ref{sec:reinforced} and show that our
new way of training allows the model to better discriminate amongst
sentences, i.e., a sentence is ranked higher for selection if it often
occurs in high scoring summaries.

\begin{table*}[t!]
  \center{\fontsize{8.5}{6.2}\selectfont 
  \begin{tabular}{| l | p{9.6cm} | c | c | c || r |}
    \hline 
    
    \multirow{7}{*}{\rotatebox[origin=c]{90}{sent. pos.}}
    & \multicolumn{1}{|c|}{\multirow{4}{*}{CNN article}}
    &
    \multirow{7}{*}{\rotatebox[origin=c]{90}{\parbox[t]{1.2cm}{Sent-level\\\textsc{rouge}}}}
    &
    \multirow{7}{*}{\rotatebox[origin=c]{90}{\parbox[t]{1.2cm}{Individual\\Oracle}}}
    &
    \multirow{7}{*}{\rotatebox[origin=c]{90}{\parbox[t]{1.2cm}{Collective\\Oracle}}}
    &
    \multicolumn{1}{|c|}{\multirow{7}{*}{\parbox[t]{1.8cm}{Multiple\\Collective\\Oracle}}}\\

    & & & & & \\ 
    & & & & & \\ 
    & & & & & \\ 
    & \multicolumn{1}{|c|}{\multirow{3}{*}{Sentences}} & & & & \\ 
    & & & & & \\
    & & & & & \\ 

    \hline  
    
    0 & A debilitating, mosquito-borne virus called Chikungunya has
    made its way to North Carolina, health officials say. & 21.2 & 1 &
    1 & \multirow{41}{*}{\parbox[t]{1.9cm}{ (0,11,13) : 59.3 \\ (0,13)
        : 57.5 \\ (11,13) : 57.2 \\ (0,1,13) : 57.1 \\ (1,13) : 56.6
        \\ (3,11,13) : 55.0 \\ (13) : 54.5 \\ (0,3,13) : 54.2
        \\ (3,13) : 53.4 \\ (1,3,13) : 52.9 \\ (1,11,13) : 52.0
        \\ (0,9,13) : 51.3 \\ (0,7,13) : 51.3 \\ (0,12,13) : 51.0
        \\ (9,11,13) : 50.4 \\ (1,9,13) : 50.1 \\ (12,13) : 49.3
        \\ (7,11,13) : 47.8 \\ (0,10,13) : 47.8 \\ (11,12,13):47.7
        \\ (7,13) : 47.6 \\ (9,13) : 47.5 \\ (1,7,13) : 46.9
        \\ (3,7,13) : 46.0 \\ (3,12,13) : 46.0 \\ (3,9,13) : 45.9
        \\ (10,13) : 45.5 \\ (4,11,13) : 45.3 \\ ... \\ \\ 
      }}\\

    1 & It's the state's first reported case of the virus. & 18.1 & 1 &
    0 & \\

    2 & The patient was likely infected in the Caribbean, according to
    the Forsyth County Department of Public Health. & 11.2 & 1 & 0 & \\

    3 & Chikungunya is primarily found in Africa, East Asia and the
    Caribbean islands, but the Centers for Disease Control and
    Prevention has been watching the virus,+ for fear that it could
    take hold in the United States -- much like West Nile did more
    than a decade ago. & 35.6 & 1 & 0 & \\

    4 & The virus, which can cause joint pain and arthritis-like
    symptoms, has been on the U.S. public health radar for some
    time. & 16.7 & 1 & 0 & \\

    5 & About 25 to 28 infected travelers bring it to the United
    States each year, said Roger Nasci, chief of the CDC's Arboviral
    Disease Branch in the Division of Vector-Borne Diseases. & 9.7 & 0
    & 0 & \\

    6 & "We haven't had any locally transmitted cases in the U.S. thus
    far," Nasci said. & 7.4 & 0 & 0 & \\

    7 & But a major outbreak in the Caribbean this year -- with more
    than 100,000 cases reported -- has health officials concerned. &
    16.4 & 1 & 0 & \\

    8 & Experts say American tourists are bringing Chikungunya back
    home, and it's just a matter of time before it starts to spread
    within the United States. & 10.6 & 0 & 0 & \\

    9 & After all, the Caribbean is a popular one with American
    tourists, and summer is fast approaching. & 13.9 & 1 & 0 & \\

    10 & "So far this year we've recorded eight travel-associated
    cases, and seven of them have come from countries in the Caribbean
    where we know the virus is being transmitted," Nasci said. & 18.4
    & 1 & 0 & \\

    11 & Other states have also reported cases of Chikungunya. & 13.4 &
    0 & 1 & \\

    12 & The Tennessee Department of Health said the state has had
    multiple cases of the virus in people who have traveled to the
    Caribbean. & 15.6 & 1 & 0 & \\
    
    13 & The virus is not deadly, but it can be painful, with symptoms
    lasting for weeks. & 54.5 & 1 & 1 & \\
    
    14 & Those with weak immune systems, such as the elderly, are more
    likely to suffer from the virus' side effects than those who are
    healthier. & 5.5 & 0 & 0 & \\

    \hline \hline

    \multicolumn{6}{|l|}{Story Highlights} \\

    \multicolumn{6}{|l|}{\parbox[t]{15cm}{\textbullet \hspace{0.1cm}
        North Carolina reports first case of mosquito-borne virus called
        Chikungunya \hspace{0.2cm} \textbullet \hspace{0.1cm}
        Chikungunya is primarily found in Africa, East Asia and the
        Caribbean islands \hspace{0.2cm} \textbullet \hspace{0.1cm}
        Virus is not deadly, but it can be painful, with symptoms
        lasting for weeks}} \\

    \hline
  \end{tabular}
  }
  \caption{An abridged CNN article (only first~15 out of~31 sentences are
    shown) and  its ``story highlights''. The
    latter are typically written by journalists to
    allow readers to quickly gather information on stories. 
    Highlights are often used as gold standard abstractive summaries 
    in the summarization literature.} \label{tab:cnnexample}
\end{table*}

\section{The Pitfalls of Cross-Entropy Loss}
\label{sec:crossent}

Previous work 
optimizes summarization
models by maximizing $p(y | \doc,\theta) = \prod_{i=1}^n p(y_i |
s_i,\doc,\theta)$, the likelihood of the ground-truth labels
\mbox{\labels = \labelseq} for sentences $(s_1,s_2,\dots,s_n)$, given
document~\doc and model parameters~$\theta$. This objective can be
achieved by minimizing the cross-entropy loss at each decoding step:
\begin{equation}
  L(\theta) = -\sum_{i=1}^n \log p(y_i | s_i,\doc,\theta). \label{eq:celoss}
\end{equation}

Cross-entropy training leads to two kinds of discrepancies in the
model. The first discrepancy comes from the disconnect between the
task definition and the training objective. While MLE in
Equation~\eqref{eq:celoss} aims to maximize the likelihood of the
ground-truth labels, the model is (a)~expected to rank sentences to
generate a summary and (b)~evaluated using $\mbox{ROUGE}$ at test
time. The second discrepancy comes from the reliance on ground-truth
labels. Document collections for training summarization systems do not
naturally contain labels indicating which sentences should be
extracted. Instead, they are typically accompanied by abstractive
summaries from which sentence-level labels are
extrapolated. \newcite{jp-acl16} follow \newcite{woodsend-acl10} in
adopting a rule-based method which assigns labels to each sentence in
the document \emph{individually} based on their semantic
correspondence with the gold summary (see the fourth column in
Table~\ref{tab:cnnexample}). An alternative method
\cite{svore-emnlp07,Cao:2016,nallapati17} identifies the set of
sentences which \emph{collectively} gives the highest ROUGE with
respect to the gold summary. Sentences in this set are labeled with~1
and 0~otherwise (see the column 5 in Table~\ref{tab:cnnexample}).

Labeling sentences individually often generates too many positive
labels causing the model to overfit the data. For example, the
document in Table~\ref{tab:cnnexample} has 12~positively labeled
sentences out of 31 in total (only first 10 are shown). Collective
labels present a better alternative since they only pertain to the few
sentences deemed most suitable to form the summary. However, a model
trained with cross-entropy loss on collective labels will underfit the
data as it will only maximize probabilities $p(1 | s_i,\doc,\theta)$
for sentences in this set (e.g., sentences $\{0,11,13\}$ in
Table~\ref{tab:cnnexample}) and ignore all other sentences. We found
that there are many candidate summaries with high ROUGE scores which
could be considered during training. 

Table~\ref{tab:cnnexample} (last column) shows candidate summaries 
ranked according to the mean of \mbox{ROUGE-1}, \mbox{ROUGE-2}, and
\mbox{ROUGE-L} F$_1$ scores. 
Interestingly, multiple top ranked summaries have reasonably high
ROUGE scores. For example, the average ROUGE for the summaries ranked
second (0,13), third (11,13), and fourth (0,1,13) is~57.5\%, 57.2\%,
and~57.1\%, and all top 16~summaries have ROUGE scores more or equal
to 50\%. A few sentences are indicative of important content and
appear frequently in the summaries: sentence~13 occurs in all
summaries except one, while sentence~0 appears in several summaries
too. Also note that summaries (11,13) and (1,13) yield better ROUGE
scores compared to longer summaries, and may be as informative, yet
more concise, alternatives.

These discrepancies render the model less efficient at ranking
sentences for the summarization task. Instead of maximizing the
likelihood of the ground-truth labels, we could train the model to
predict the individual ROUGE score for each sentence in the document
and then select the top $m$~sentences with highest scores. But
sentences with individual ROUGE scores do not necessarily lead to a
high scoring summary, e.g.,~they may convey overlapping content and
form verbose and redundant summaries. For example, sentence~3, despite
having a high individual ROUGE score (35.6\%), does not occur in any
of the top~5 summaries. We next explain how we address these issues
using reinforcement learning.

\section{Sentence Ranking with Reinforcement Learning} 
\label{sec:reinforced}

Reinforcement learning \cite{Sutton98a} has been proposed as a way of
training sequence-to-sequence generation models in order to directly
optimize the metric used at test time, e.g.,~BLEU or ROUGE
\cite{ranzato-arxiv15-bias}. We adapt reinforcement learning to our
formulation of extractive summarization to rank sentences for summary
generation.
We propose an objective function that combines the maximum-likelihood
cross-entropy loss with rewards from policy gradient reinforcement
learning to globally optimize ROUGE. Our training algorithm allows to
explore the space of possible summaries, making our model more robust
to unseen data. As a result, reinforcement learning helps extractive
summarization in two ways: (a)~it directly optimizes the evaluation
metric instead of maximizing the likelihood of the ground-truth labels
and (b)~it makes our model better at discriminating among sentences; a
sentence is ranked high for selection if it often occurs in high
scoring summaries.


\subsection{Policy Learning}

We cast the neural summarization model introduced in
Figure~\ref{fig:architecture} in the Reinforcement Learning paradigm
\cite{Sutton98a}. Accordingly, the model can be viewed as an ``agent''
which interacts with an ``environment'' consisting of documents. At
first, the agent is initialized randomly, it reads document~\doc and
predicts a relevance score for each sentence $s_i \in D$ using
``policy'' $p(y_i | s_i,\doc,\theta)$, where $\theta$~are model
parameters. Once the agent is done reading the document, a summary
with labels $\hat{y}$ is sampled out of the ranked sentences.  The
agent is then given a ``reward''~$r$ commensurate with how well the
extract resembles the gold-standard summary. Specifically, as reward
function we use mean F$_1$ of $\mbox{ROUGE-1}$, $\mbox{ROUGE-2}$,
and $\mbox{ROUGE-L}$. Unigram and bigram overlap ($\mbox{ROUGE-1}$ and
$\mbox{ROUGE-2}$) are meant to assess informativeness, whereas the
longest common subsequence ($\mbox{ROUGE-L}$) is meant to assess
fluency.
We update the agent using the REINFORCE algorithm \cite{Williams:1992}
which aims to minimize the negative expected reward:
\begin{equation}
  L(\theta) = -\mathbb{E}_{\hat{y} \sim\ p_{\theta}}
  [r(\hat{y})] \label{eq:reinloss}
\end{equation}
where, $p_{\theta}$ stands for $p(y|\doc,\theta)$. REINFORCE is based
on the observation that the expected gradient of a non-differentiable
reward function (ROUGE, in our case) can be computed as
follows:
%
\begin{equation}
 \nabla L(\theta) = -\mathbb{E}_{\hat{y} \sim\ p_{\theta}} [r(\hat{y})
   \nabla \log p(\hat{y} | \doc,\theta)] \label{eq:reinlossgrad}
\end{equation}
While MLE in Equation~\eqref{eq:celoss} aims to maximize the
likelihood of the training data, the objective in
Equation~\eqref{eq:reinloss} learns to discriminate among sentences
with respect to how often they occur in high scoring summaries.

\subsection{Training with High Probability Samples}

Computing the expectation term in Equation~\eqref{eq:reinlossgrad} is
prohibitive, since there is a large number of possible extracts. In
practice, we approximate the expected gradient using a single
sample~$\hat{y}$ from~$p_{\theta}$ for each training example in a
batch:
\begin{align}
  \nabla L(\theta) &\approx -r(\hat{y}) \nabla \log
  p(\hat{y} | \doc,\theta) \\ &\approx - r(\hat{y}) \sum_{i=1}^n \nabla
  \log p(\hat{y_i} | s_i,\doc,\theta) \label{eq:reinlossgrad-singsample}
\end{align}

Presented in its original form, the REINFORCE algorithm starts
learning with a random policy which can make model training
challenging for complex tasks like ours where a single document can
give rise to a very large number of candidate summaries. We therefore
limit the search space of~$\hat{y}$ in
Equation~\eqref{eq:reinlossgrad-singsample} to the set of largest
probability samples~$\hat{\mathbb{Y}}$. We
approximate~$\hat{\mathbb{Y}}$ by the~$k$ extracts which receive
highest ROUGE scores. More concretely, we assemble candidate summaries
efficiently by first selecting $p$~sentences from the document which
on their own have high ROUGE scores. We then generate all possible
combinations of $p$~sentences subject to maximum length $m$ and
evaluate them against the gold summary. Summaries are ranked according
to~F$_1$ by taking the mean of $\mbox{ROUGE-1}$, $\mbox{ROUGE-2}$, and
$\mbox{ROUGE-L}$. $\hat{\mathbb{Y}}$ contains these top~$k$ candidate
summaries.  During training, we sample~$\hat{y}$
from~$\hat{\mathbb{Y}}$ instead of~$p(\hat{y} | \doc,\theta)$.

\newcite{ranzato-arxiv15-bias} proposed an alternative to REINFORCE
called MIXER (Mixed Incremental Cross-Entropy Reinforce) which first
pretrains the model with the cross-entropy loss using ground truth
labels and then follows a curriculum learning strategy
\cite{bengio-nips2015-curriculum} to gradually teach the model to
produce stable predictions on its own. In our experiments MIXER
performed worse than the model of \newcite{nallapati17} just trained
on collective labels. We conjecture that this is due to the unbounded
nature of our ranking problem. Recall that our model assigns relevance
scores to sentences rather than words. The space of sentential
representations is vast and fairly unconstrained compared to other
prediction tasks operating with fixed vocabularies
\cite{li-emnlp-16,paulus-socher-arxiv17,xingxing-arxiv-17}.  Moreover,
our approximation of the gradient allows the model to converge much
faster to an optimal policy. Advantageously, we do not require an
online reward estimator, we pre-compute $\hat{\mathbb{Y}}$, which
leads to a significant speedup during training compared to MIXER
\cite{ranzato-arxiv15-bias} and related training schemes
\cite{shenMRT-acl16}.




\section{Experimental Setup}
\label{sec:experiments}

In this section we present our experimental setup for assessing the
performance of our model which we call \refresh\ as a shorthand for
\textbf{RE}in\textbf{F}o\textbf{R}cement Learning-based
\textbf{E}xtractive \textbf{S}ummarization. We describe our datasets,
discuss implementation details, our evaluation protocol, and the
systems used for comparison.

\paragraph{Summarization Datasets}

We evaluated our models on the CNN and DailyMail news highlights
datasets \cite{hermann-nips15}. We used the standard splits of
\newcite{hermann-nips15} for training, validation, and testing
(90,266/1,220/1,093 documents for CNN and 196,961/12,148/10,397 for
DailyMail).  We did not anonymize entities or lower case tokens. We
followed previous studies
\cite{jp-acl16,nallapati-signll16,nallapati17,see-acl17,tanwan-acl17}
in assuming that the ``story highlights'' associated with each article
are gold-standard abstractive summaries.  During training we use these
to generate high scoring extracts and to estimate rewards for them,
but during testing, they are used as reference summaries to evaluate
our models.



\paragraph{Implementation Details}

We generated extracts by selecting three sentences ($m=3$) for CNN
articles and four sentences ($m=4$) for DailyMail articles. These
decisions were informed by the fact that gold highlights in the
CNN/DailyMail validation sets are 2.6/4.2~sentences long.  For both
datasets, we estimated high-scoring extracts using 10~document
sentences ($p=10$) with highest ROUGE scores. We tuned the
initialization parameter $k$~for $\hat{\mathbb{Y}}$ on the validation
set: we found that our model performs best with $k=5$~for the CNN
dataset and $k=15$~for the DailyMail dataset.

We used the One Billion Word Benchmark corpus \cite{billionbenchmark}
to train word embeddings with the skip-gram model \cite{word2vec}
using context window size~6, negative sampling size~10, and
hierarchical softmax~1.  Known words were initialized with pre-trained
embeddings of size~200. Embeddings for unknown words were initialized
to zero, but estimated during training. Sentences were padded with
zeros to a length of~100. For the sentence encoder, we used a list of
kernels of widths 1 to 7, each with output channel size of 50
\cite{kim-aaai16}. The sentence embedding size in our model was~350.

For the recurrent neural network component in the document encoder and
sentence extractor, we used a single-layered LSTM network with
size~600. All input documents were padded with zeros to a maximum
document length of~120. We performed minibatch cross-entropy training
with a batch size of 20~documents for 20~training epochs. It took
around 12 hrs on a single GPU to train. After each epoch, we evaluated
our model on the validation set and chose the best performing model
for the test set. During training we used the Adam optimizer
\cite{adam-14} with initial learning rate~$0.001$. Our system is
implemented in TensorFlow \cite{tensorflow2015-whitepaper}.

\begin{figure}[t!]

  \center{\fontsize{8.5}{6.2}\selectfont 
    \begin{tabular}{|@{~}c@{~}| p{6.8cm} |}
      \hline 
      \multirow{10}{*}{\rotatebox[origin=c]{90}{\textsc{Lead}}} & 

      \textbullet \hspace{0.1cm} A SkyWest Airlines flight made an
      emergency landing in Buffalo, New York, on Wednesday after a
      passenger lost consciousness, officials said. \\

      & \textbullet \hspace{0.1cm} The passenger received medical
      attention before being released, according to Marissa Snow,
      spokeswoman for SkyWest. \\

      & \textbullet \hspace{0.1cm} She said the airliner expects to
      accommodate the 75 passengers on another aircraft to their
      original destination -- Hartford, Connecticut -- later Wednesday
      afternoon. \\ \hline

      \multirow{6}{*}{\rotatebox[origin=c]{90}{See et al.}} &

      \textbullet \hspace{0.1cm} Skywest Airlines flight made an
      emergency landing in Buffalo, New York, on Wednesday after a
      passenger lost consciousness. \\

      & \textbullet \hspace{0.1cm} She said the airliner expects to
      accommodate the 75 passengers on another aircraft to their
      original destination -- Hartford, Connecticut. \\ \hline
   
      \multirow{8}{*}{\rotatebox[origin=c]{90}{\refresh}} &

      \textbullet \hspace{0.1cm} A SkyWest Airlines flight made an
      emergency landing in Buffalo, New York, on Wednesday after a
      passenger lost consciousness, officials said. \\

      & \textbullet \hspace{0.1cm} The passenger received medical
      attention before being released, according to Marissa Snow,
      spokeswoman for SkyWest. \\

      & \textbullet \hspace{0.1cm} The Federal Aviation Administration
      initially reported a pressurization problem and said it would
      investigate. \\ \hline
      
      \multirow{4}{*}{\rotatebox[origin=c]{90}{\textsc{Gold}}} 
      
      & \textbullet \hspace{0.1cm} FAA backtracks on saying crew reported a pressurization problem \\ 

      & \textbullet \hspace{0.1cm} One passenger lost consciousness \\ 

      & \textbullet \hspace{0.1cm} The plane descended 28,000 feet in three minutes \\ \hline \hline
      
      Q$_1$ & Who backtracked on saying crew reported a pressurization problem? (\emph{FAA}) \\ 
      Q$_2$ & How many passengers lost consciousness in the incident? (\emph{One}) \\
      Q$_3$ & How far did the plane descend in three minutes? (\emph{28,000 feet}) \\ \hline
    \end{tabular}
  }
  \caption{Summaries produced by the \textsc{Lead} baseline, the
    abstractive system of \protect\newcite{see-acl17} and \refresh\ for
    a CNN (test) article. \textsc{Gold} presents the human-authored
    summary; the bottom block shows manually written questions using
    the gold summary and their answers in
    parentheses.}\label{fig:summaries}
  \vspace{-0.2cm}
\end{figure}

\paragraph{Evaluation}
We evaluated summarization quality using F$_1$ $\mbox{ROUGE}$
\cite{rouge}. We report unigram and bigram overlap ($\mbox{ROUGE-1}$
and $\mbox{ROUGE-2}$) as a means of assessing informativeness and the
longest common subsequence ($\mbox{ROUGE-L}$) as a means of assessing
fluency.\footnote{We used pyrouge, a Python package, to compute all
  ROUGE scores with parameters ``-a -c 95 -m -n 4 -w 1.2.''}  We
compared \refresh\ against a baseline which simply selects the first
$m$~leading sentences from each document (\textsc{Lead}) and two
neural models similar to ours (see left block in
Figure~\ref{fig:architecture}), both trained with cross-entropy loss.
\newcite{jp-acl16} train on individual labels, while
\newcite{nallapati17} use collective labels. We also compared our
model against the abstractive systems of \newcite{chenIjcai-16},
\newcite{nallapati-signll16}, \newcite{see-acl17}, and
\newcite{tanwan-acl17}.\footnote{\newcite{jp-acl16} report ROUGE
  recall scores on the DailyMail dataset only. We used their code
  (\url{https://github.com/cheng6076/NeuralSum}) to produce ROUGE
  F$_1$ scores on both CNN and DailyMail datasets. For other systems,
  all results are taken from their papers.}

In addition to ROUGE which can be misleading when used as the only
means to assess the informativeness of summaries
\cite{schluter:2017:EACLshort}, we also evaluated system output by
eliciting human judgments in two ways. In our first experiment,
participants were presented with a news article and summaries
generated by three systems: the \textsc{Lead} baseline, abstracts from
\newcite{see-acl17}, and extracts from \refresh. We also included the
human-authored highlights.\footnote{We are grateful to Abigail See for
  providing us with the output of her system. We did not include
  output from \newcite{nallapati17}, \newcite{chenIjcai-16},
  \newcite{nallapati-signll16}, or \newcite{tanwan-acl17} in our human
  evaluation study, as these models are trained on a named-entity
  anonymized version of the CNN and DailyMail datasets, and as result
  produce  summaries which are not comparable to ours. We
  did not include extracts from \newcite{jp-acl16} either as they were
  significantly inferior to \textsc{Lead} (see
  Table~\ref{tab:cnndm}).}  Participants read the articles and were
asked to rank the summaries from best (1) to worst (4) in order of
informativeness (does the summary capture important information in the
article?) and fluency (is the summary written in well-formed
English?). We did not allow any ties. We randomly selected 10~articles
from the CNN test set and 10~from the DailyMail test set. The study
was completed by five participants, all native or proficient English
speakers. Each participant was presented with the 20 articles. The
order of summaries to rank was randomized per article and the order of
articles per participant. Examples of summaries our subjects ranked
are shown in Figure~\ref{fig:summaries}.

Our second experiment assessed the degree to which our model retains
key information from the document following a question-answering (QA)
paradigm which has been previously used to evaluate summary quality
and text compression \cite{MorrisKA92,Mani:1999,Clarke:Lapata:2010}.
We created a set of questions based on the gold summary under the
assumption that it highlights the most important document content. We
then examined whether participants were able to answer these questions
by reading system summaries alone without access to the article. The
more questions a system can answer, the better it is at summarizing
the document as a whole.


We worked on the same 20 documents used in our first elicitation
study. We wrote multiple fact-based question-answer pairs for each
gold summary without looking at the document. Questions were
formulated so as to not reveal answers to subsequent questions.  We
created 71~questions in total varying from two to six questions per
gold summary. Example questions are given in
Figure~\ref{fig:summaries}. Participants read the summary and answered
all associated questions as best they could without access to the
original document or the gold summary.  Subjects were shown summaries
from three systems: the \textsc{Lead} baseline, the abstractive system
of \newcite{see-acl17}, and \refresh. Five participants answered
questions for each summary. We used the same scoring mechanism from
\newcite{Clarke:Lapata:2010}, i.e., a correct answer was marked with a
score of one, partially correct answers with a score of~0.5, and zero
otherwise. The final score for a system is the average of all its
question scores. Answers were elicited using Amazon's Mechanical Turk
crowdsourcing platform. We uploaded data in batches (one system at a
time) on Mechanical Turk to ensure that same participant does not
evaluate summaries from different systems on the same set of
questions.

\begin{table*}[t!]
  \center{\small
  \begin{tabular}{  |l |c c c|  c c c|  c c c|  }
    \hline 

\multicolumn{1}{|c|}{Models}&    \multicolumn{3}{c}{CNN} &
    \multicolumn{3}{c}{DailyMail} &
    \multicolumn{3}{c|}{CNN$+$DailyMail}\\ 

     & R1 & R2 & RL & R1 & R2 & RL & R1 & R2 & RL \\ \hline \hline 
    
    \textsc{Lead} (ours) & 29.1 & 11.1 & 25.9 & 40.7 & 18.3 & 37.2 &
    39.6 & 17.7 & 36.2 \\
    
    \textsc{Lead}$^{\ast}$ \cite{nallapati17} & --- & --- & --- & --- & --- & --- & 39.2  & 15.7 & 35.5 \\
    
    \textsc{Lead} \cite{see-acl17} & --- & --- & --- & --- & --- & --- & \textbf{40.3} & 17.7 & \textbf{36.6} \\
    
    \newcite{jp-acl16} & 28.4 & 10.0 & 25.0 & 36.2 & 15.2 & 32.9 & 35.5 & 14.7 & 32.2 \\ 
    
    
    \newcite{nallapati17}$^{\ast}$ & --- & --- & --- & --- & --- & --- & 39.6 & 16.2 & 35.3 \\ 

    \refresh\ & \textbf{30.4} & \textbf{11.7}  & \textbf{26.9}  & \textbf{41.0}  & \textbf{18.8}  & \textbf{37.7}  & 40.0  & \textbf{18.2}  & \textbf{36.6}  \\ \hline \hline 
        
    \newcite{chenIjcai-16}$^{\ast}$ & 27.1 & 8.2 & 18.7 & --- & --- & --- & --- & --- & --- \\

    \newcite{nallapati-signll16}$^{\ast}$ & --- & --- & --- & --- & --- & --- & 35.4 & 13.3 & 32.6 \\ 
    
\newcite{see-acl17} & --- & --- & --- & --- & --- & --- & 39.5 & 17.3 & 36.4 \\
    
    
    \newcite{tanwan-acl17}$^{\ast}$ & 30.3 & 9.8 & 20.0 & --- & --- & --- & 38.1 & 13.9 & 34.0 \\   \hline
  \end{tabular}
  }
  \caption{Results on the CNN and DailyMail test sets. We report
    ROUGE-1 (R1), ROUGE-2 (R2), and ROUGE-L (RL) F$_1$
    scores. Extractive systems are in the first block and
    abstractive systems in the second.  Table cells are filled with ---
    whenever results are not available.  Models marked with $^{\ast}$
    are not directly comparable to ours as they are based on
    an anonymized version of the dataset.\label{tab:cnndm}} 
  \vspace{-0.2cm}
\end{table*}

\section{Results} 
\label{sec:results}

We report results using automatic metrics in
Table~\ref{tab:cnndm}. The top part of the table compares \refresh\
against related extractive systems. The bottom part reports the
performance of abstractive systems. We present three variants of
\textsc{Lead}, one is computed by ourselves and the other two are
reported in \newcite{nallapati17} and \newcite{see-acl17}. Note that
they vary slightly due to differences in the preprocessing of the
data. We report results on the CNN and DailyMail datasets and their
combination (CNN$+$DailyMail).


\paragraph{Cross-Entropy vs Reinforcement Learning}

The results in Table~\ref{tab:cnndm} show that \refresh\ is superior
to our \textsc{Lead} baseline and extractive systems across datasets
and metrics. It outperforms the extractive system of
\newcite{jp-acl16} which is trained on individual labels.  \refresh\
is not directly comparable with \newcite{nallapati17} as they generate
anonymized summaries. Their system lags behind their \textsc{Lead}
baseline on ROUGE-L on the CNN+DailyMail dataset (35.5\% vs 35.3\%).
Also note that their model is trained on collective labels and has a
significant lead over \newcite{jp-acl16}. As discussed in
Section~\ref{sec:crossent} cross-entropy training on individual labels
tends to overgenerate positive labels leading to less informative and
verbose summaries.


%

\paragraph{Extractive vs Abstractive Systems}

Our automatic evaluation results further demonstrate that
\refresh\ is superior to abstractive systems
\cite{chenIjcai-16,nallapati-signll16,see-acl17,tanwan-acl17} which
are all variants of an encoder-decoder architecture
\cite{sutskever-nips14}. Despite being more faithful to the actual
summarization task (hand-written summaries combine several pieces of
information from the original document), abstractive systems lag
behind the \textsc{Lead} baseline. \newcite{tanwan-acl17} present a
graph-based neural model, which manages to outperform \textsc{Lead} on
ROUGE-1 but falters when higher order ROUGE scores are used. Amongst
abstractive systems \newcite{see-acl17} perform best. Interestingly,
their system is mostly extractive, exhibiting a small degree of
rewriting; it copies more than 35\% of the sentences in the source
document, 85\% of 4-grams, 90\% of 3-grams, 95\% of bigrams, and 99\%
of unigrams.

\begin{table}[t!]
  \center{\footnotesize
    \begin{tabular}{|@{~}l@{~~}| c@{~~} c@{~~} c@{~~} c@{~~} |c@{~}| } 
      \hline
      Models & 1st & 2nd & 3rd & 4th & QA\\ \hline
      \textsc{Lead} & 0.11 & 0.21 & \textbf{0.34} & {0.33} & 36.33\\ 
      \newcite{see-acl17} & 0.14 & 0.18 & 0.31 & \textbf{0.36} & 28.73 \\
      \refresh\ & 0.35 &\textbf{0.42} & 0.16 & 0.07 & \textbf{66.34} \\ 
     \textsc{Gold} & \textbf{0.39} & 0.19 & 0.18 & 0.24 & ---\\ 
      \hline
    \end{tabular}}
  \caption{System ranking and QA-based evaluations. Rankings (1st,
    2nd, 3rd and 4th) are shown as proportions. Rank 1 is the best and
    Rank 4, the worst. The column QA shows the percentage of questions
    that participants answered correctly by reading system
    summaries. \label{tab:heval}}
  \vspace{-0.2cm}
\end{table}

\paragraph{Human Evaluation: System Ranking}



Table~\ref{tab:heval} shows, proportionally, how often participants
ranked each system, 1st, 2nd, and so on. Perhaps unsurprisingly
human-authored summaries are considered best (and ranked 1st 39\% of
the time). \refresh\ is ranked 2nd best followed by \textsc{Lead} and
\newcite{see-acl17} which are mostly ranked in 3rd and 4th places. We
carried out pairwise comparisons between all models in
Table~\ref{tab:heval} to assess whether system differences are
statistically significant. There is no significant difference between
\textsc{Lead} and \newcite{see-acl17}, and \refresh\ and \textsc{Gold}
(using a one-way ANOVA with post-hoc Tukey HSD tests; \mbox{$p <
  0.01$}). All other differences are statistically significant.

\paragraph{Human Evaluation: Question Answering}



The results of our QA evaluation are shown in the last column of
Table~\ref{tab:heval}. Based on summaries generated by \refresh,
participants can answer 66.34\% of questions correctly. Summaries
produced by \textsc{Lead} and the abstractive system of
\newcite{see-acl17} provide answers for 36.33\% and 28.73\% of the
questions, respectively. Differences between systems are all
statistically significant (\mbox{$p < 0.01$}) with the exception of
\textsc{Lead} and \newcite{see-acl17}.

Although the QA results in Table~\ref{tab:heval} follow the same
pattern as ROUGE in Table~\ref{tab:cnndm}, differences among systems
are now greatly amplified. QA-based evaluation is more focused and a
closer reflection of users' information need (i.e.,~to find out what
the article is about), whereas ROUGE simply captures surface
similarity (i.e.,~\mbox{$n$-gram} overlap) between output summaries
and their references.  Interestingly, \textsc{Lead} is considered
better than \newcite{see-acl17} in the QA evaluation, whereas we find
the opposite when participants are asked to rank systems. We
hypothesize that \textsc{Lead} is indeed more informative than
\newcite{see-acl17} but humans prefer shorter summaries. The average
length of \textsc{Lead} summaries is 105.7~words compared to~61.6 for
\newcite{see-acl17}.



\section{Related Work}

Traditional summarization methods manually define features to rank
sentences for their salience in order to identify the most important
sentences in a document or set of documents
\cite{Kupiec:1995binary,mani2001automatic,radev-lrec2004,filatova-04event,nenkova-06,SparckJones:2007}.
A vast majority of these methods learn to score each sentence
independently
\cite{Barzilay97usinglexical,Teufel97sentenceextraction,erkan:2004:lexrank,Mihalcea04TextRank,Shen:2007:IJCAI,Schilder:2008:fastsum,Wan:2010:urank}
and a summary is generated by selecting top-scored sentences in a way
that is not incorporated into the learning process. Summary quality
can be improved heuristically, \cite{Yih:2007:MSM}, via max-margin
methods \cite{Carbonell:1998:UMD,Li:2009:EDC}, or integer-linear
programming
\cite{woodsend-acl10,berg:2011,Woodsend:2012:emnlp,Almeida:acl13,parveen:2015:tgraph}. 

Recent deep learning methods
\cite{krageback-cvsc14,Yin-ijcai15,jp-acl16,nallapati17} learn
continuous features without any linguistic preprocessing (e.g., named
entities). Like traditional methods, these approaches also suffer from
the mismatch between the learning objective and the evaluation
criterion (e.g., ROUGE) used at the test time.  In comparison, our
neural model globally optimizes the ROUGE evaluation metric through a
reinforcement learning objective: sentences are highly ranked if they
occur in highly scoring summaries.


Reinforcement learning has been previously used in the context of
traditional multi-document summarization as a means of selecting a
sentence or a subset of sentences from a document
cluster. \newcite{Ryang:2012} cast the sentence selection task as a
search problem. Their agent observes a state (e.g.,~a candidate
summary), executes an action (a transition operation that produces a
new state selecting a not-yet-selected sentence), and then receives a
delayed reward based on $\mbox{tf}*\mbox{idf}$. Follow-on work
\cite{Rioux:emnlp14} extends this approach by employing ROUGE as part
of the reward function, while \newcite{hens-gscl15} further experiment
with \mbox{$Q$-learning}. \newcite{MollaAliod:2017:ALTA2017} has adapt
this approach to query-focused summarization. Our model differs from
these approaches both in application and formulation. We focus solely
on extractive summarization, in our case states are documents (not
summaries) and actions are relevance scores which lead to sentence
ranking (not sentence-to-sentence transitions). Rather than employing
reinforcement learning for sentence selection, our algorithm performs
sentence ranking using ROUGE as the reward function.


The REINFORCE algorithm \cite{Williams:1992} has been shown to improve
encoder-decoder text-rewriting systems by allowing to directly
optimize a non-differentiable objective
\cite{ranzato-arxiv15-bias,li-emnlp-16,paulus-socher-arxiv17} or to
inject task-specific constraints
\cite{xingxing-arxiv-17,nogueira-cho:2017:EMNLP2017}. However, we are
not aware of any attempts to use reinforcement learning for training a
sentence ranker in the context of extractive summarization. 







\section{Conclusions}
\label{sec:conclusions}

In this work we developed an extractive summarization model which is
globally trained by optimizing the ROUGE evaluation metric. Our
training algorithm explores the space of candidate summaries while
learning to optimize a reward function which is relevant for the task
at hand. Experimental results show that reinforcement learning offers
a great means to steer our model towards generating informative,
fluent, and concise summaries outperforming state-of-the-art
extractive and abstractive systems on the CNN and DailyMail
datasets. In the future we would like to focus on smaller discourse
units \cite{rst} rather than individual sentences, modeling
compression and extraction
jointly. 

\paragraph{Acknowledgments} \begin{footnotesize} We gratefully
  acknowledge the support of the European Research Council (Lapata;
  award number 681760), the European Union under the Horizon 2020
  SUMMA project (Narayan, Cohen; grant agreement 688139), and Huawei
  Technologies (Cohen).  The research is based upon work supported by
  the Office of the Director of National Intelligence (ODNI),
  Intelligence Advanced Research Projects Activity (IARPA), via
  contract FA8650-17-C-9118.  The views and conclusions contained
  herein are those of the authors and should not be interpreted as
  necessarily representing the official policies or endorsements,
  either expressed or implied, of the ODNI, IARPA, or the
  U.S. Government. The U.S. Government is authorized to reproduce and
  distribute reprints for Governmental purposes notwithstanding any
  copyright annotation thereon.\end{footnotesize}

\bibliography{summarisation-improved.bib}
\bibliographystyle{acl_natbib} 

\end{document}